\documentclass[sigconf]{acmart}
\usepackage{makecell}
\usepackage{graphicx} %use graph format
\usepackage{multirow}
\usepackage{tablefootnote}
\usepackage{threeparttable}
\usepackage{subcaption}
\usepackage{balance}
\usepackage{booktabs}
\acmSubmissionID{XXX}

\begin{document}

%\title{Multimodal Emotion Recognition in Noisy Environment Based on Progressive Label Revision }
\title[Temporal Label Hierachical Network for Compound Emotion Recognition]{Temporal Label Hierachical Network for Compound Emotion Recognition}
\author{Sunan Li$^\dag$}
\affiliation{%
\institution{School of Information Science and Engineering, Southeast University}
\streetaddress{1 Th{\o}rv{\"a}ld Circle}
\city{Nanjing}
\country{China}}
\email{230189473@seu.edu.cn}

\author{Hailun Lian$^\dag$}
\affiliation{%
\institution{School of Information Science and Engineering, Southeast University}
\streetaddress{1 Th{\o}rv{\"a}ld Circle}
\city{Nanjing}
\country{China}}
\email{lianhailun@seu.edu.cn}

\author{Cheng Lu}
\affiliation{%
\institution{School of Biological Science and Medical Engineering, Southeast University}
\streetaddress{1 Th{\o}rv{\"a}ld Circle}
\city{Nanjing}
\country{China}}
\email{cheng.lu@seu.edu.cn}

\author{Yan Zhao}
\affiliation{%
\institution{School of Information Science and Engineering, Southeast University}
\streetaddress{1 Th{\o}rv{\"a}ld Circle}
\city{Nanjing}
\country{China}}
\email{zhaoyan@seu.edu.cn}

\author{Tianhua Qi}
\affiliation{%
\institution{School of Biological Science and Medical Engineering, Southeast University}
\streetaddress{1 Th{\o}rv{\"a}ld Circle}
\city{Nanjing}
\country{China}}
\email{}

\author{Hao Yang}
\affiliation{%
\institution{School of Information Science and Engineering, Southeast University}
\streetaddress{1 Th{\o}rv{\"a}ld Circle}
\city{Nanjing}
\country{China}}
\email{}

\author{Yuan Zong$^*$}
\affiliation{%
\institution{School of Biological Science and Medical Engineering, Southeast University}
\streetaddress{1 Th{\o}rv{\"a}ld Circle}
\city{Nanjing}
\country{China}}
\email{xhzongyuan@seu.edu.cn}

\author{Wenming Zheng$^*$}
\affiliation{%
\institution{Key Laboratory of Child Development and Learning Science of Ministry of Education, School of Biological Science and Medical Engineering, Southeast University}
\streetaddress{1 Th{\o}rv{\"a}ld Circle}
\city{Nanjing}
\country{China}}
\email{wenming\_zheng@seu.edu.cn}

\thanks{$^{*}$Corresponding author.}
\thanks{$^\dag$Both authors contributed equally to this research. }
\renewcommand{\shortauthors}{}
\begin{abstract}
The emotion recognition has attracted more attention in recent decades. Although significant progress has been made in the recognition technology of the seven basic emotions, existing methods are still hard to tackle compound emotion recognition that occurred commonly in practical application. This article introduces our achievements in the 7th Field Emotion Behavior Analysis (ABAW) competition. In the competition, we selected pre trained ResNet18 and Transformer, which have been widely validated, as the basic network framework. Considering the continuity of emotions over time, we propose a time pyramid structure network for frame level emotion prediction. Furthermore. At the same time, in order to address the lack of data in composite emotion recognition, we utilized fine-grained labels from the DFEW database to construct training data for emotion categories in competitions. Taking into account the characteristics of valence arousal of various complex emotions, we constructed a classification framework from coarse to fine in the label space.
\end{abstract}

\begin{CCSXML}

\end{CCSXML}

\ccsdesc[500]{Human-centered computing~Human computer interaction (HCI)}
\ccsdesc[500]{Computing methodologies~Artificial intelligence}

\keywords{Multimodal emotion recognition, compound emotion, feature fusion}


\maketitle

\section{INTRODUCTION}
Emotion recognition is a technology aimed at endowing machines with the ability to identify, process, and understand human emotions.  For example, previous work often using elaborated designed hand-crafted features such as LBP and IS09 and machine learning based methods support vector machine (SVM) \cite{BHAVAN2019104886}, Gaussian mixture model (GMM) \cite{2007GMM}, supervised dictionary learning \cite{2014Multiview} and sparse representation \cite{MENCATTINI201468} \cite{2016Speech} to classify emotion class. In recent years, with the rapid advancement of deep learning techniques, various emotion recognition methods have been proposed. For example, in \cite{li2023multimodal} Li et.al using label revision method to cope with emotion recognition in nosiy environments. In \cite{lu2022implicitly} Lu et.al impose sparse constrain on the reconstruction matrix to select more effective features

However, these methods often focus on the recognition of seven basic emotions. In practical applications, complex emotions composed of combinations of these basic emotions are more commonly encountered. There is relatively little research in this domain, and the lack of high-quality databases for complex emotions hinders further development in this field. To facilitate the development of compound emotions recognition, the 7th Affective Behavior
Analysis in-the-wild (ABAW) hold the Compound Expression (CE) Recogntion based on the C-EXPR-DB database\cite{kollias20246th,kollias2023abaw2,kollias2023multi,kollias2023abaw,kollias2022abaw,kollias2021analysing,kollias2021affect,kollias2021distribution,kollias2020analysing,kollias2019expression,kollias2019deep,kollias2019face,zafeiriou2017aff,kollias20247th}.

The remainder of this paper is organized as follows. The framework of our modal including training dataset preparation and the backbone of our methods are described in section 2. In section 3, we show our experimental results on the challenge dataset to evaluate the effectiveness of our proposed method. Finally, in section 4, we conclude this paper.

\section{THE PROPOSED METHOD}

\subsection{Feature Extraction and Fusion}
We first employed the OpenFace toolkit for face detection in every frame of the video. For images where faces were difficult to detect, we supplemented with the closest temporally adjacent face to obtain a face image corresponding to every original video frame. Considering the temporal continuity of emotional states, despite the task requiring emotion classification prediction for each frame, we constructed a temporal pyramid structure of image sequences to acquire more robust emotional features. Three sets of image sequences at different temporal scales were composed as follows: a sequence of 15 frames starting from the current frame, a sequence of 15 frames sampled from a quarter-length segment of the video where the current frame resides, and a sequence of 15 frames sampled from the entire video. These hierarchical sequences of three image sets were parallelly fed into a spatiotemporal feature extraction network consisting of ResNet18 and Transformer. For each frame image, we averaged the sum of all classification results obtained from different image sequences to derive the final classification result.

Additionally, due to inherent data imbalance in the training set, as depicted in Fig. 1 of emotional cycles, we utilized the DFEW database to train the network for positive and negative classification in valence and arousal to assist the final classification results. Specifically, for each frame, if both valence and arousal were positive, it was directly categorized as a compound emotion of happiness-surprise; if both were negative, a judgment was made among three compound emotions carrying sadness components. Since other sets of compound emotions do not exhibit mutual exclusivity in valence and arousal, valence and arousal were not used as assisting information for their recognition. The overall network framework is illustrated in Fig. 2.

\begin{figure*}[t!]
\centering
\begin{subfigure}[b]{0.6\textwidth}
\includegraphics[width=\textwidth]{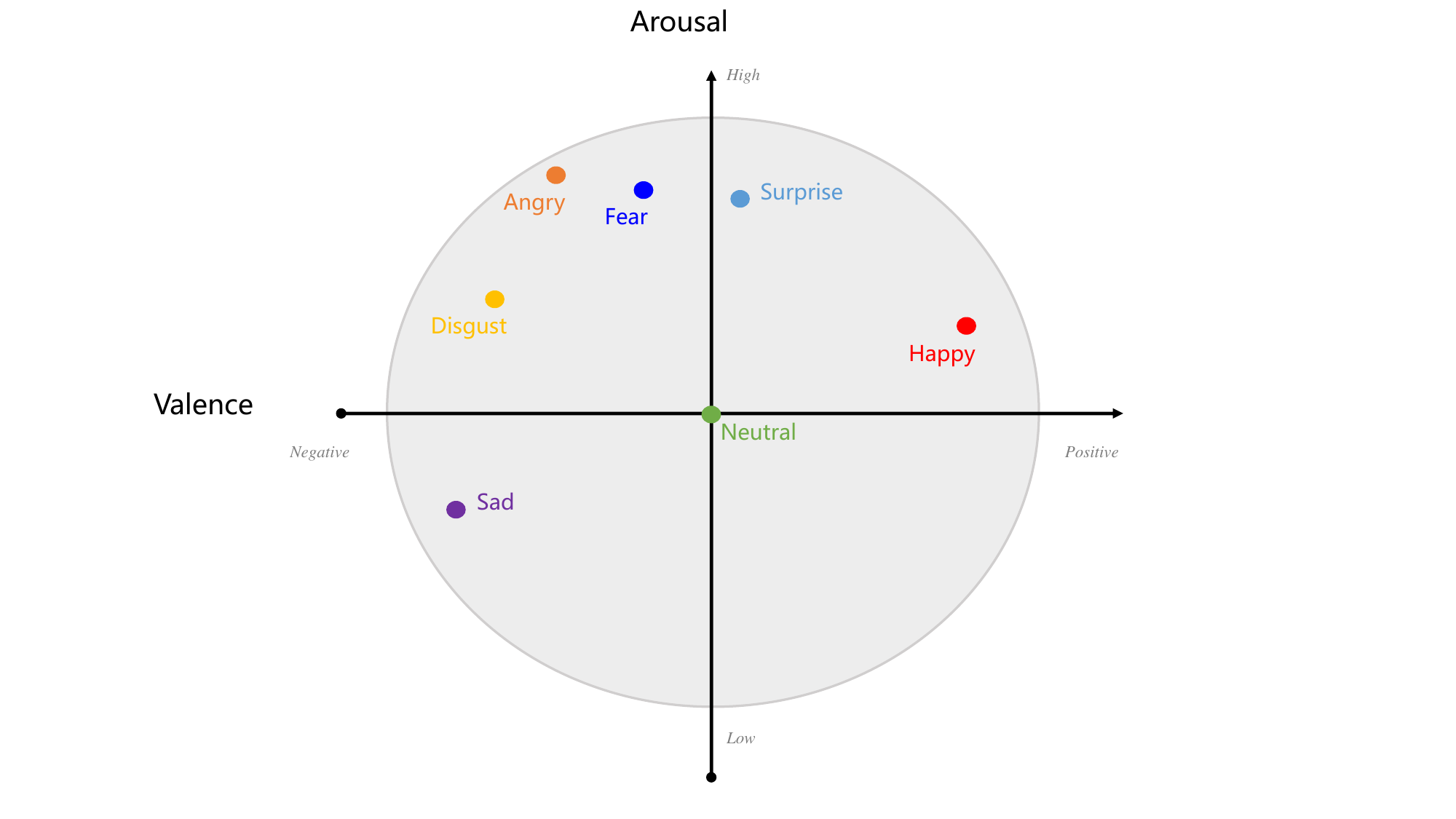}
\caption{The emotion cycle.}
\label{fig_1}
\end{subfigure}

\end{figure*}

\begin{figure*}[t!]
\centering
\begin{subfigure}[b]{0.9\textwidth}
\includegraphics[width=\textwidth]{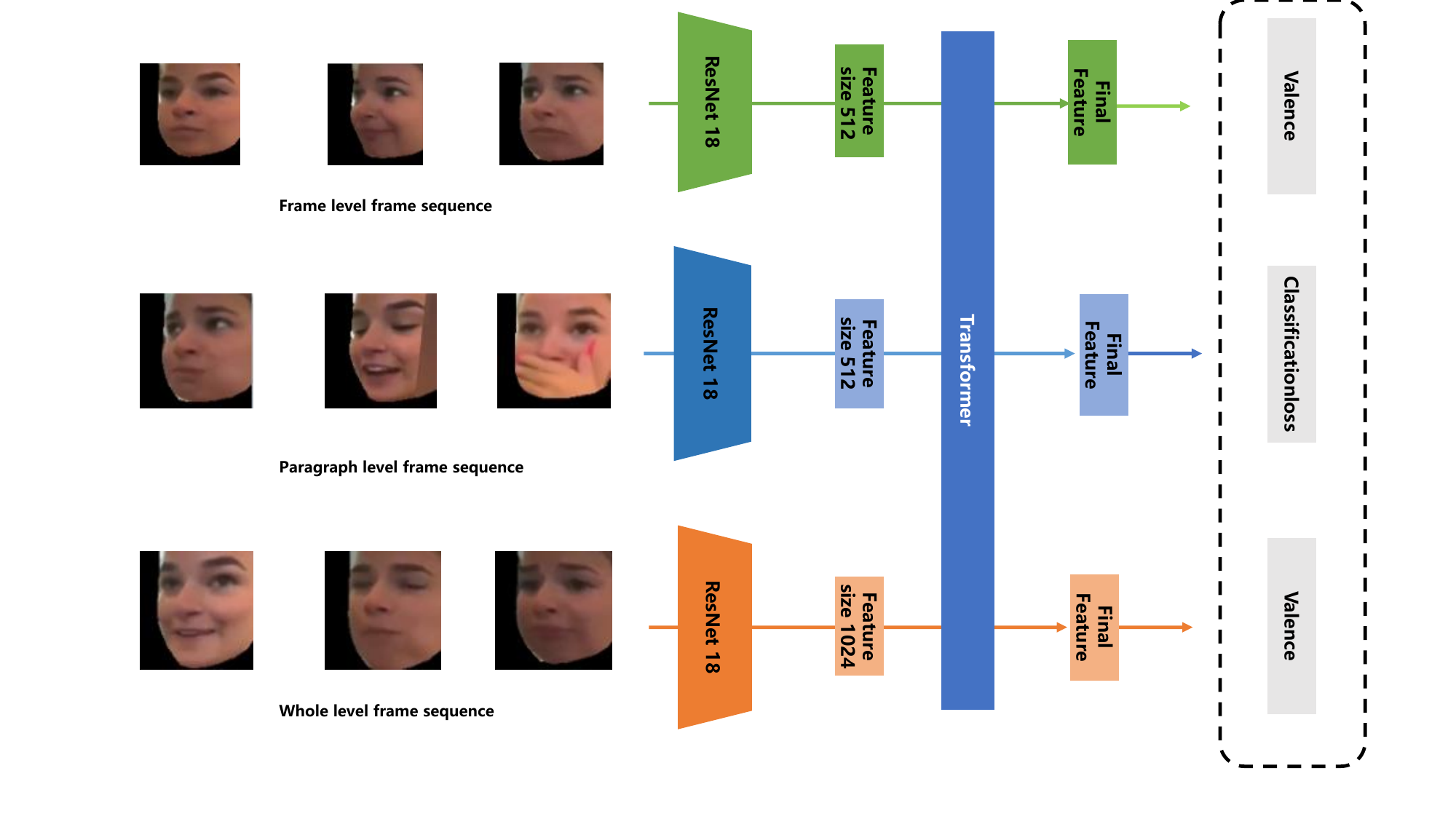}
\caption{The overall structure of the network.}
\label{fig_2}
\end{subfigure}

\end{figure*}

\section{EXPERIMENT}

\subsection{Dataset and Preprocessing}

To select appropriate training and validation data, we utilized the DFEW database, which features detailed emotion labels. Established by Jiang et al. in 2020, this database initially collected over 1500 high-resolution movie clips depicting near-real scenarios, yielding 16,372 facial expression videos. Each video segment was independently labeled by 10 annotators with one of the basic emotions (happiness, sadness, neutral, anger, surprise, disgust, fear). The final true label for each video segment was determined based on emotions chosen by more than 6 annotators. Ultimately, 12,059 video segments were selected. For the competition task, we curated a training set comprising 1864 samples, ensuring each component of composite emotions was represented by ratings from at least 3 annotators. Considering significant sample imbalances and the mutual exclusivity of happiness and disgust across the seven composite emotions, additional single-emotion data from the DFEW database were included to balance the dataset. Furthermore, recognizing the unique positions of happiness and sadness on the emotional wheel, we performed valence and arousal-based positive classification using the DFEW database to assist in determining the final emotional categories.

\subsection{Training settings}
All image resolution used in this paper is consistently set to 224 × 224. During training, the number of epochs is set to 50. Cross-entropy is utilized as the classification loss function, and the Adam is selected as the optimizer, and the learning rate is set at 3e-4 according to experiment performance, and the batch size is 90.

\subsection{Result and Discussion}

According to the performance assessment rules of the competition, the evaluation the performance of compound expressions recognition by the average F1 Score across all 7 compound expressions. Therefore, the evaluation criterion is:

\section{CONCLUSION}
In this paper, we propose a hierarchical composite emotion recognition network in both temporal and label spaces. The emotional category for each frame is determined by aggregating classification information from image sequences across different time spans to provide final discriminative information. Simultaneously, a hierarchical classification strategy is designed based on the differences in emotional dynamics across emotional composites. The final results demonstrate promising performance on ABAW7.

\bibliographystyle{ACM-Reference-Format}
\balance
\bibliography{MERNOISE}

\end{document}